\documentclass[a4paper,fleqn]{cas-dc}

\usepackage[numbers]{natbib}

\usepackage{float}
\usepackage{algorithm}
\usepackage{algpseudocode}

\def\tsc#1{\csdef{#1}{\textsc{\lowercase{#1}}\xspace}}
\tsc{WGM}
\tsc{QE}
\tsc{EP}
\tsc{PMS}
\tsc{BEC}
\tsc{DE}

\begin{document}
\let\WriteBookmarks\relax
\def\floatpagepagefraction{1}
\def\textpagefraction{.001}
\shorttitle{Chatlaw: multi-agent legal assistant}
\shortauthors{Jiaxi Cui et~al.}

\fntext[fn1]{These authors contribute equally to this article.}
\cortext[cor1]{Corresponding author.}

\author[a,d]{Jiaxi Cui}[orcid=0009-0001-2291-559X]
\fnmark[1]

\author[a,b]{Munan Ning}[orcid=0009-0005-3418-085X]
\fnmark[1]

\author[a]{Zongjian Li}[]
\fnmark[1]

\author[a,b]{Hao Li}[]
\fnmark[1]

\author[a]{Yang Yan}[]
\author[a]{Bohua Chen}[]
\author[c]{Bin Ling}[]
\author[a,b]{Yonghong Tian}[]

\author[a,b]{Li Yuan}[orcid=0000-0002-2120-5588]
\cormark[1]
\ead{yuanli-ece@pku.edu.cn}

\affiliation[a]{organization={Shenzhen Graduate School, Peking University},
                city={Shenzhen},
                country={China}}

\affiliation[b]{organization={Peng Cheng Laboratory},
                city={Shenzhen},
                country={China}}

\affiliation[c]{organization={Law School, Peking University},
                city={Beijing},
                country={China}}

\affiliation[d]{organization={Pandalla.ai},
                city={Beijing},
                country={China}}

\title [mode = title]{Chatlaw: A Multi-Agent Legal Assistant based on a Role-Aligned Mixture-of-Experts Architecture}
\tnotemark[1]
\tnotetext[1]{\copyright{} 2026 The Authors. This manuscript version is made available under the CC BY-NC-ND 4.0 license. The Version of Record is available at \url{https://doi.org/10.1016/j.fmre.2026.03.026}.}

\begin{abstract}
Artificial Intelligence (AI) holds great potential in legal services, yet Large Language Models (LLMs) face two major challenges: limited knowledge of the Chinese legal system and vulnerability to hallucinations. To address these issues, we present Chatlaw, a multi-agent legal assistant. Chatlaw's framework is designed to emulate the Standard Operating Procedures (SOP) of real law firms, where different roles (e.g., assistant, researcher, senior lawyer) collaborate on a case. To computationally mirror this collaborative structure, we developed a novel Role-Aligned Mixture-of-Experts (RA-MoE) architecture. In this system, the internal "experts" are specifically trained to align with the distinct tasks of each agent role (e.g., inquiry, analysis, drafting). These specialized agents (Legal Assistant, Researcher, etc.) then form the collaborative framework. When they interact with users, retrieve legal knowledge, analyze case details, or generate reliable consultations, the RA-MoE architecture intelligently routes their computations to the corresponding dedicated expert, ensuring each step is handled by the most qualified parameters. In evaluations, Chatlaw surpasses general-purpose AI models, including GPT-4, achieving a 7.73\% improvement in accuracy on the LawBench benchmark and an 11-point higher score on the Unified Qualification Exam for Legal Professionals. Real-case studies and expert assessments further confirm its robustness. Chatlaw enhances the accessibility and reliability of legal services, advancing the provision of legal support to the public.
\end{abstract}

\begin{keywords}
artificial intelligence in law \sep mixture of experts \sep large language model \sep knowledge graph \sep legal assistant \sep legal technology
\end{keywords}

\maketitle

\section{Introduction}


Legal services play a crucial role in protecting individual rights and maintaining social fairness~\cite{heinz2005urban,auerbach1977unequal,tushnet2009rights}. However, the limited availability of legal professionals and the high cost of their services often restrict access to these services, particularly in China, with its vast population and extensive social interactions. This gap in legal service provision deeply impacts justice and equity, especially for those lacking the resources to effectively navigate the legal system. 
Besides the widespread demand for legal services, China's legal system has two significant characteristics: First, unlike the United States, where each state has different legal regulations, China's legal system is unified. Second, China's centralized legal system and extensive legal infrastructure provide a unique and suitable environment for the research and development of legal AI. Therefore, this raises a crucial question: can we establish an automated legal assistant to address the specific needs of one of the world's largest and most uniform legal systems?


In recent years, the efficacy of Large Language Models (LLMs) has been validated across multiple scientific fields, encompassing natural language processing~\cite{brown2020language,touvron2023llama2}, human interaction~\cite{kirk2024benefits,strachan2024testing}, biochemistry~\cite{ren2023multiplexed,lv2024prollama,peng2024large,kuenneth2023polybert,theodorou2023synthesize,suvarna2023language,choi2024accelerating,ferruz2022protgpt2}, and the medical field~\cite{thirunavukarasu2023large,clusmann2023future,gilbert2023large,singh2023augmenting,fan2024deep,kiyasseh2024framework,nigo2024deep,tayebi2024large}. LLMs also offer potential solutions to the challenges in legal services. Popular models like ChatGPT~\cite{openai2023gpt4} and the LLaMA~\cite{touvron2023llama} series, along with other general-purpose~\cite{bai2023qwen,du2021glm,zeng2022glm,2023internlm,cai2024internlm2,yang2023baichuan} or law-specific models,~\cite{nguyen2023brief,lawyer-llama-report,deng2023syllogistic} can respond to user inputs based on their internal legal knowledge repositories and provide advisory recommendations. However, the inherent \textit{hallucination} issues in LLMs pose potential risks in their application to legal domains~\cite{mik2024caveat} since they operate at the level of word distributions rather than validated facts~\cite{mccoy2023embers}.
The knowledge generated by these models is often incomplete or outdated, leading them to produce illusions that, although seemingly relevant, may be misleading or incorrect~\cite{mik2024caveat,maynez2020faithfulness}.

To address these limitations, we propose Chatlaw, a multi-agent virtual legal assistant based on a novel architecture designed to align computational specialization with real-world legal workflows. The core innovation lies in the Role-Aligned Mixture-of-Experts (RA-MoE) architecture. Rather than using a single LLM for all tasks, we leverage a task-specific Mixture-of-Experts (MoE) model~\cite{li2024mixlora,hua2025input}, where each "expert" is specialized for specific tasks based on the Standard Operating Procedures (SOPs) of legal service workflows. SOPs in law firms guide task delegation (e.g., information gathering, research, legal advice, report writing), and this framework ensures that each agent's role is matched with an optimal, task-specific model~\cite{boiko2023autonomous}.

This alignment of experts to specific legal functions mitigates hallucination risks and enhances both efficiency and accuracy. For example, a model optimized for legal research may not be ideal for generating legal advice or drafting documents. By tailoring each agent to its function, we improve both task-specific performance and overall workflow efficiency. Additionally, we support this system with a high-quality legal dataset and knowledge graph, ensuring the accuracy of the information provided by each agent.

Our model demonstrates superior performance in various evaluations, surpassing existing large language models, including GPT-4~\cite{openai2023gpt4}. Specifically, it outperforms GPT-4 in the Lawbench and Unified Qualification Exam for Legal Professionals
by 7.73\% in accuracy and 11 points, respectively, and receives the highest scores in real-case evaluation feedback from legal experts on 4 dimensions, completeness, correctness, guidance, and authority.

\section{Related Work}
\label{sec:related_work}
\subsection{Mixture-of-Experts Architecture}
MoE is a powerful technique used to achieve computational efficiency in large models. The core idea of MoE is to replace dense feed-forward networks with a sparse mixture of "expert" layers, where only a subset of experts is activated during each inference step. This allows models to scale more effectively by significantly reducing the computational cost, especially in terms of floating-point operations (FLOPs). Early work on MoE includes GShard \cite{lepikhin2020gshard}, which demonstrated the scalability of MoE to large-scale systems, and Switch Transformers \cite{fedus2022switch}, which improved performance while reducing computational redundancy by activating a sparse subset of experts. More recently, open-source models like DeepSeek~\cite{liu2024deepseek} and Mixtral 8x7B \cite{jiang2024mixtral} have demonstrated the potential of MoE in delivering state-of-the-art performance in natural language processing tasks.

Despite its successes, traditional MoE architectures suffer from a lack of task-specific specialization. In standard MoE models, the router learns to assign tokens to experts based on the distribution of data~\cite{krajewski2024scaling}, without aligning these assignments with any specific high-level task such as reasoning or drafting. This general-purpose specialization limits MoE's ability to handle domain-specific tasks where expert alignment with specific roles is essential. To address this limitation, our work introduces a RA-MoE architecture, where the router is explicitly trained to align expert selections with high-level agent roles defined in our multi-agent framework. This ensures that the experts activated in each task are functionally specialized, thereby enhancing the system's performance in complex multi-agent workflows.

\subsection{Multi-Agent Systems}
Multi-agent systems (MAS)~\cite{hong2023metagpt} have been explored as a means to overcome the limitations of single monolithic models in complex task execution. In these systems, multiple agents with specialized functions collaborate to achieve a common goal. Early frameworks, such as ReAct \cite{yao2022react}, enabled LLMs to call external tools by combining reasoning and acting capabilities. Building on this, more advanced collaborative frameworks like AutoGen \cite{wu2024autogen} have been developed, where multiple agents with different specializations collaborate to solve complex problems.

However, most existing multi-agent systems focus primarily on the macro-level orchestration of workflows~\cite{pei2025flow,gan2025master}. These systems still rely on the same underlying, general-purpose LLM APIs (e.g., GPT-4) for all agents, leading to significant computational redundancy. This issue arises because all agents, regardless of their specialized roles, are forced to share the same computational resources.

Our work addresses this challenge by not only defining the agent-based SOP at the macro level, but also introducing a novel micro-level architecture, the RA-MoE. This architecture ensures that experts are specifically aligned with the roles of the agents, thereby improving computational efficiency and reducing redundancy in multi-agent systems.

\subsection{AI in Law}
The application of artificial intelligence in the legal domain has a long history, with early work focusing on legal information retrieval and rule-based expert systems \cite{niebla2018artificial,gao2023retrieval}. These systems aimed to automate basic legal tasks such as legal research and document classification. The advent of LLMs has significantly advanced AI in Law, with recent efforts focusing on enhancing the depth of legal knowledge and reasoning capabilities of AI systems.

Legal benchmarks such as LawBench \cite{fei2023lawbench} and LexGLUE \cite{chalkidis2021lexglue} have been developed to evaluate and improve the performance of LLMs in the legal domain. In addition, domain-specific large models like Lawyer LLaMA \cite{huang2023lawyer}, LawGPT \cite{zhou2024lawgpt} and Fuzi-Mingcha~\cite{sdu_fuzi_mingcha} have been created by fine-tuning general-purpose LLMs on legal corpora. While these models have achieved impressive results, they still face significant challenges when handling complex legal consultations. These challenges include functional generalization, where a single model is expected to proficiently handle diverse tasks such as legal retrieval, reasoning, and drafting, and reliability, where the monolithic nature of a single model makes it prone to hallucinations in long-chain tasks \cite{dhuliawala2023chain}. Our work aims to address these challenges by constructing a multi-agent framework that mirrors the SOP of a law firm, where different roles collaborate to ensure the accuracy and consistency of legal outputs.

\begin{figure*}[]
  \centering
  \includegraphics[width=0.9\textwidth]{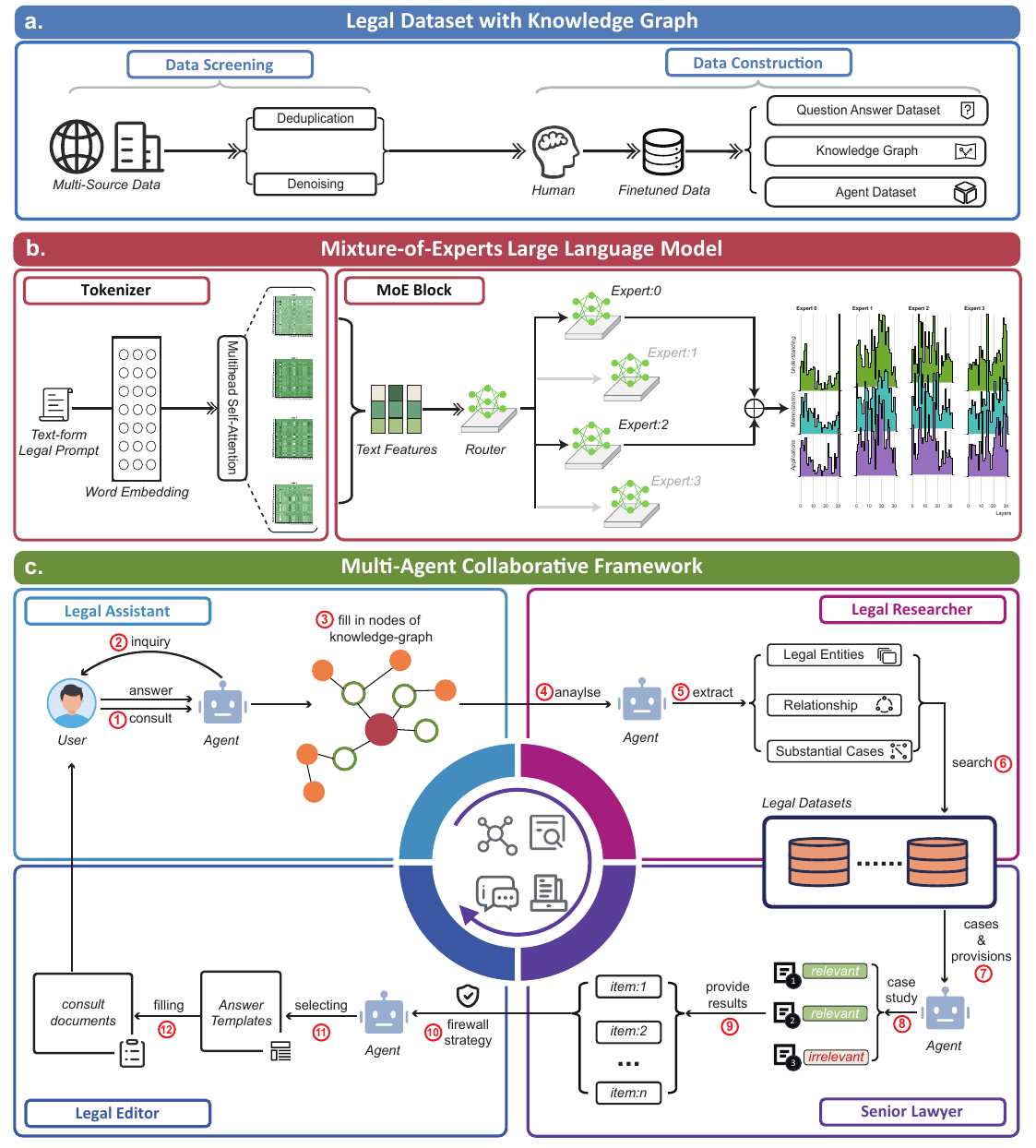}
  \caption{\textbf{The Framework of Chatlaw.} 
    Our framework consists of three core components:
    \textbf{(a) Data Processing Pipeline:} We screen and structure multi-source data to construct a high-quality legal dataset enriched with a knowledge graph.
    \textbf{(b) Mixture-of-Experts (MoE) Model:} The core language model utilizes an MoE architecture, where a router directs legal prompts to specialized expert sub-networks for efficient and accurate processing.
    \textbf{(c) Multi-Agent Collaborative Framework:} A system of four specialized agents (\textbf{Legal Assistant}, \textbf{Researcher}, \textbf{Editor}, and \textbf{Senior Lawyer}) that simulates a legal team's workflow. The framework handles the entire consultation process, from initial user inquiry and legal research to document editing and final review.}
  \label{fig:main}
\end{figure*}

\section{Chatlaw Framework}

\subsection{Role-Aligned Mixture-of-Experts}

In this work, we introduce the RA-MoE framework, which extends traditional MoE models by introducing role-conditioned routing and knowledge-aware gating. Traditional MoE architectures, while efficient in terms of computational cost, typically rely on sparse expert activation without considering the semantic diversity of tasks or the context of knowledge required for specific tasks. In contrast, RA-MoE strategically aligns expert activation with the role-specific tasks in legal workflows and integrates domain-specific knowledge into the routing mechanism to reduce errors and hallucinations during legal consultations.

\subsubsection{Role-Conditioned Routing}
A key innovation of RA-MoE is the introduction of role-conditioned routing. In traditional MoE models, expert selection is based purely on token-level representations, which are agnostic to the task or the role performing the task. However, in the context of legal services, different stages of legal consultation require distinct expert specializations. For instance, legal research, document drafting, and high-level legal reasoning each demand different parameter sets. In RA-MoE, the router input consists of three components: (i) the token embedding \( x \), representing the content of the input text, (ii) the role embedding \( E_a \), which represents the current agent's role in the workflow (e.g., Legal Assistant, Legal Researcher, Legal Editor, Senior Lawyer), and (iii) the knowledge-graph embedding \( E_k \), encoding the relevant legal entities and relationships extracted from a structured legal knowledge base.

The routing distribution over \( K \) experts is computed as:
\[
G(x, E_a, E_k; \tau) = \text{softmax} \left( \frac{W_g [x; E_a; E_k]}{\tau} \right),
\]
where \( \tau \) controls the sharpness of the distribution. The top \( K \) experts are selected based on the highest probabilities, and their outputs are aggregated as a weighted sum:
\[
\text{RA-MoE}(x, E_a, E_k) = \sum_{i=1}^{K} G_i(x, E_a, E_k) \, \mathcal{E}_i(x),
\]
where \( \mathcal{E}_i \) denotes the \( i \)-th expert's feed-forward sub-network. This role-conditioned routing ensures that the appropriate experts are activated for each legal task, leading to more accurate and interpretable results.

\subsubsection{Knowledge-Aware Expert Activation}
RA-MoE further incorporates knowledge-aware gating to enhance its ability to handle legal information effectively. The knowledge-graph embedding \( E_k \) is crucial for reducing hallucinations and ensuring the reliability of the output. By embedding structured legal knowledge into the routing mechanism, we enable the system to activate experts who are specialized in handling specific legal entities (e.g., laws, cases, statutes) and their relationships. This integration is particularly useful in legal domains where the accuracy of facts is paramount, and hallucinations can lead to legal risks. The knowledge-aware gating allows RA-MoE to filter out irrelevant expert activations and focus on experts that are most relevant to the legal domain context, thus improving overall system reliability.

\subsubsection{Agent–Expert Co-Training}
In addition to the role-conditioned routing and knowledge-aware gating, RA-MoE introduces agent–expert co-training. Unlike static MoE models where experts are trained independently, RA-MoE allows for shared training between agents and their corresponding expert clusters. During the multi-agent collaboration process, the Legal Assistant, Legal Researcher, Legal Editor, and Senior Lawyer agents leverage expert clusters specific to their tasks, but they also share gradients across tasks to ensure coherence throughout the entire consultation process. This ensures that the system operates efficiently, with each agent invoking experts dynamically based on the legal task at hand, while maintaining the overall integrity of the legal consultation pipeline.

The co-training process is guided by a task-expert routing loss function \( \mathcal{L}_{route} \), which encourages the router to activate the correct expert subset based on the role-specific task. The total loss function for training is:
\[
\mathcal{L}_{total} = \mathcal{L}_{reg} + \alpha \mathcal{L}_{route},
\]
where \( \mathcal{L}_{reg} \) is the standard cross-entropy loss for autoregressive training, and \( \mathcal{L}_{route} \) penalizes deviations between the routing probabilities and the intended expert selection.

The \( \mathcal{L}_{route} \) loss function is defined as:
\[
\mathcal{L}_{route} = - \sum_{e=1}^{E} T(e | t) \log G_e(x, E_a, E_k; \tau),
\]
where \( T(e | t) \) is the target distribution for expert \( e \) given task \( t \), \( G_e(x, E_a, E_k; \tau) \) is the routing probability output for expert \( e \), and \( \tau \) is the temperature parameter controlling the sharpness of the distribution. This loss function helps ensure that each task type activates the most appropriate experts, improving both efficiency and accuracy.

\begin{algorithm}[H]
\caption{Generic Agent Template in the SOP Framework}
\label{alg:agent_template}
\begin{algorithmic}[1]
\Require Role identifier $r$, Expert cluster $\mathcal{E}_r$, 
Task configuration $\mathcal{P}_r$, Input $I_t$ (from user or previous agent)
\Ensure Structured output $O_t$

\State \textbf{Initialization:} 
Load role embedding $E_a(r)$ and task prompt $\mathcal{P}_r$ for role $r$
\State \textbf{Input Reception:} 
Receive $I_t$ from user (if first agent) or previous agent output $O_{t-1}$
\State \textbf{Pre-Processing:} 
Parse $I_t$ into structured context $C_t$ (entities, keywords, graph nodes)
\State \textbf{Expert Invocation (via RA-MoE):}
\Statex \hspace{1em}Compute routing distribution 
$G(x,E_a,E_k) = \mathrm{softmax}(W_g [x;E_a;E_k]/\tau)$
\Statex \hspace{1em}Activate top-$K$ experts $\mathcal{E}_r$ aligned with role $r$
\Statex \hspace{1em}Generate intermediate output 
$Y_t = \sum_{i=1}^{K} G_i(x,E_a,E_k)\,\mathcal{E}_i(x)$
\State \textbf{Post-Processing:} 
Validate and refine $Y_t$ through factual verification and formatting
\State \textbf{Output Transmission:} 
Store refined output $O_t$ in shared memory; 
pass $O_t$ to the next agent in the SOP sequence
\end{algorithmic}
\end{algorithm}

\subsection{Chatlaw Agent Design}

Built upon the RA-MoE architecture, Chatlaw establishes a multi-agent SOP framework that reproduces the workflow of a real-world law firm.
Each agent represents a distinct professional role (including the Legal Assistant, Legal Researcher, Legal Editor, and Senior Lawyer) and operates according to a unified execution template.
This structure divides a complex legal consultation into four coordinated stages, where each stage is handled by experts optimized for its specific function.

Within this framework, every agent follows a generic operational template.
The template specifies how the agent receives inputs, activates its corresponding expert cluster through the RA-MoE, processes information, and delivers structured outputs to the next stage of the workflow.
Through this unified design, the system maintains both functional specialization (achieved by role-aligned experts) and procedural coherence (achieved by standardized task execution).
Algorithm \ref{alg:agent_template} outlines the general procedure followed by all agents in the SOP framework.

Following the unified template, the four agents perform complementary functions that collectively realize the complete legal consultation workflow.
The Legal Assistant serves as the entry point of the system and mirrors the role of a paralegal or junior associate in a law firm.
It receives unstructured user inquiries as input and, through the Pre-Processing stage of the template, extracts entities, temporal information, and key issues to build a structured knowledge graph.
This graph acts as a factual foundation that is transmitted to the next stage.
The Legal Researcher, corresponding to a legal analyst, then processes this structured graph as its input.
Through the Expert Invocation procedure of the template, it activates the retrieval-oriented experts within the RA-MoE to search external databases and statutory repositories for relevant laws, judicial interpretations, and precedent cases.
The output of this stage is a curated corpus of verified legal materials that grounds the subsequent reasoning process.

The following stage is handled by the Senior Lawyer, who represents an experienced legal advisor.
The Senior Lawyer takes the verified corpus and the structured knowledge graph as input and performs high-level legal reasoning using the advisory experts of the RA-MoE.
It validates the document’s factual integrity and provides the final professional opinion of the SOP pipeline.
Finally, the Legal Editor takes the Senior Lawyer’s output as input and engages the document-generation experts to compose a well-structured draft of the legal report or consultation document.
Through the Post-Processing phase of the template, the Legal Editor ensures stylistic consistency and formal compliance before passing the draft onward to the user.
Together, these four agents embody the modular workflow of a real law firm, where each stage corresponds to a distinct professional responsibility and collectively ensures that the consultation is both accurate and procedurally coherent.

Table~\ref{tab:agent_expert} summarizes the configuration of each agent and its corresponding expert cluster, along with the core responsibilities and Input/Output behaviors.

\begin{table*}[]
\centering
\caption{Role–Expert Mapping and Functional Responsibilities in the SOP Framework}
\label{tab:agent_expert}
\begin{tabular}{llll}
\hline
\textbf{Agent Role} & \textbf{Expert Cluster} & \textbf{Primary Function} & \textbf{Input / Output} \\
\hline
Legal Assistant & $\mathcal{E}_{1..2}$ & Entity extraction and graph building & Input: query $\rightarrow$ Output: structured graph \\[3pt]
Legal Researcher & $\mathcal{E}_{3..4}$ & Retrieval of laws and cases & Input: graph $\rightarrow$ Output: corpus \\[3pt]
Senior Lawyer & $\mathcal{E}_{5..6}$ & Legal reasoning and validation & Input: draft $\rightarrow$ Output: opinion \\[3pt]
Legal Editor & $\mathcal{E}_{7..8}$ & Drafting and formatting legal docs & Input: corpus $\rightarrow$ Output: draft \\[3pt]

\hline
\end{tabular}
\end{table*}

\subsection{SOP Integration and Workflow Execution}
The complete SOP integrates the four agents into a sequential workflow that mirrors the collaboration pattern of a professional legal team.
The process begins with the Legal Assistant, who interacts directly with the user and converts raw inquiries into a structured representation through knowledge-graph construction.
This structured information is then transferred to the Legal Researcher, which performs retrieval and evidence gathering, grounding the subsequent reasoning on verifiable legal sources.
The refined corpus is passed to the Legal Editor, which consolidates the retrieved knowledge and user context into standardized legal documents such as consultation reports or petitions.
Finally, the Senior Lawyer reviews the draft, validates the legal reasoning, and formulates the authoritative opinion that completes the consultation cycle.

Throughout the workflow, all intermediate results are stored in a shared latent memory that maintains continuity across stages.
When each agent executes its task, the RA-MoE router dynamically activates the expert cluster aligned with the agent’s role, allowing specialized computation while preserving global context.
This design ensures that information flows transparently between agents without redundancy or loss, enabling the system to emulate the procedural rigor of an actual law firm, where assistants prepare data, researchers verify facts, editors formalize content, and senior lawyers deliver the final decision.
By coupling RA-MoE-based specialization with SOP-driven coordination, Chatlaw achieves an interpretable and reliable framework for multi-stage legal reasoning and documentation.

\begin{table*}[htpb]
\centering
\caption{Legal Data Used for Training the Large Legal Model.}
\label{tab:legal_data_restructured}
\begin{tabular}{@{} l c c c @{}}
    \toprule
    \textbf{Sub-Category} & \textbf{Sample Size} & \textbf{{Scope of Legal Coverage}} & \textbf{{Temporal Validity}} \\
    \midrule
    
    \multicolumn{4}{l}{\textbf{Statutes}} \\ 
    \quad Legal provisions & 23k & {Civil, Criminal, Admin, etc.} & {Updated to 2025.11}\\ 
    \quad Legal interpretations & 71k & {SPC \& SPP Interpretations} & {Updated to 2025.11}\\
    \quad Regulations & 35k & {National \& Local Regulations} & {Updated to 2025.11}\\
    \midrule 
        
    \multicolumn{4}{l}{\textbf{Cleaned Cases}} \\
    \quad Precedents & 796k & {National Court Hierarchy} & {2000--2025}\\
    \quad Typical cases & 59k & {SPC Guiding \& Typical Cases} & {2010--2025}\\
    \quad Cases of public interest & 665k & {Civil \& Admin Public Interests} & {2015--2025}\\
    \midrule
        
    \multicolumn{4}{l}{\textbf{Legal AI Competition Data}} \\
    \quad LAIC & 768k & {Legal Knowledge Reasoning} & {2018--2024}\\
    \quad CAIL & 312k & {Judgment \& Statute Prediction} & {2018--2022}\\
    \quad Other Competitions & 220k & {Diverse Legal NLP Tasks} & {2018--2024}\\
    \midrule
        
    \multicolumn{4}{l}{\textbf{Legal Examination}} \\
    \quad Examination questions & 25k & {National Qualification Exam} & {2018--2024}\\
    \quad Examination answers & 25k & {Official \& Model Answers} & {2018--2024}\\
    \quad Examination analysis & 38k & {Expert Reasoning \& Analysis} & {2018--2024}\\
    \midrule
        
    \multicolumn{4}{l}{\textbf{Consultation + Legal Q\&A}} \\
    \quad Lawyer consultation & 73k & {Professional Case Consultations} & {2020--2025}\\
    \quad Public consultation & 242k & {Community Legal Queries} & {2020--2025}\\
    \quad Legal Q\&A & 42k & {General Legal Knowledge Base} & {2020--2025}\\
    \bottomrule
\end{tabular}
\end{table*}

\begin{table}[htpb]
\centering
\caption{Task Data Used for Training the Large Legal Model.}
\label{tab:task_data_grouped}
\begin{tabular}{@{} p{5cm} @{\hspace{-35pt}} r  r @{}}

    \toprule
    \textbf{Sub-Category} & \textbf{Sample Size} & \textbf{{Improving Agents}} \\
    \midrule
    
    \multicolumn{3}{l}{\textbf{Text Processing}} \\
    \quad Keyword extraction & 52k & {Legal Researcher} \\
    \quad Text classification & 73k & {Legal Assistant} \\
    \quad Text generation & 70k & {Legal Editor} \\
    \quad Issue Topic Identification & 18k & {Legal Assistant} \\
    \quad Reading Comprehension & 29k & {Senior Lawyer} \\
    \midrule
    
    \multicolumn{3}{l}{\textbf{Data Annotation}} \\
    \quad Named Entity Recognition & 24k & {Legal Assistant} \\ 
    \quad Relationship extraction & 18k & {Legal Assistant} \\
    \quad Intent recognition & 21k & {Legal Assistant} \\
    \quad Dispute Focus Recognition & 21k & {Legal Assistant} \\
    \quad Argument Mining & 22k & {Senior Lawyer} \\
    \midrule

    \multicolumn{3}{l}{\textbf{Document Processing}} \\
    \quad Document proofreading & 19k & {Legal Editor} \\
    \quad Document merging & 6k & {Legal Editor} \\
    \quad Document splitting & 32k & {Legal Editor} \\
    \midrule

    \multicolumn{3}{l}{\textbf{Legal Process Simulation}} \\
    \quad Sub-case segmentation & 42k & {Senior Lawyer} \\
    \quad Clarify details & 31k & {Legal Assistant} \\
    \quad Legal document drafting & 5k & {Legal Editor} \\
    \quad Legal argumentation & 21k & {Senior Lawyer} \\
    \quad Event Detection & 35k & {Legal Assistant} \\
    \quad Trigger Word Extraction & 28k & {Legal Researcher} \\
    \quad Statute Prediction & 13k & {Legal Researcher} \\
    \quad Charge Prediction & 12k & {Senior Lawyer} \\
    \quad Sentence Prediction & 28k & {Senior Lawyer} \\
    \midrule            

    \multicolumn{3}{l}{\textbf{Simulated Judgment}} \\
    \quad Statute matching & 120k & {Legal Researcher} \\
    \quad Sentence calculation & 73k & {Senior Lawyer} \\
    \quad Inference of charges & 48k & {Senior Lawyer} \\
    \quad Criminal Amount Calculation & 91k & {Senior Lawyer} \\
    \bottomrule
\end{tabular}
\end{table}

\subsection{Comprehensive Legal Data Pipeline and Preprocessing for Multi-Task Training}

To enable Chatlaw to efficiently handle multi-stage legal tasks, we have developed a comprehensive data pipeline that integrates both legal dataset construction and preprocessing.  
This pipeline is designed to generate high-quality task-oriented datasets and ensure they are preprocessed into a format suitable for large language model training.  
As shown in Fig.~\ref{fig:main}a, the process begins with {multi-source data collection}, gathering data from statutes, case documents, legal competition datasets, public consultations, and legal examinations.  
We employ automated tools for {deduplication, denoising, and standardization} to ensure the data is cleaned and converted into a consistent format, making it suitable for large-scale model training.  
The result is a diverse legal dataset, as detailed in Table~\ref{tab:legal_data_restructured}, with over 3.4 million samples covering multiple legal domains, {with detailed scope of legal coverage and temporal validity}.

After the data is collected, we implement a {hybrid human-machine curation process} to enhance the quality and accuracy of the data.  
Law students participate in labeling cases and categorizing them, while domain experts define the semantic relationships between legal terms and create structured knowledge graphs.  
This process ensures the data is not only comprehensive but also legally valid and interpretable, providing the foundation for specialized tasks in the Chatlaw framework.  
Table~\ref{tab:task_data_grouped} shows the task-oriented datasets {with specified types of agents associated with each sub-category during the training phase}, which include 26 distinct tasks such as keyword extraction, statute prediction, document drafting, and simulated judgment. 
{The training data for RA-MoE is not explicitly divided into subsets for separate agent training; rather, the model is trained on the full dataset as a unified system.}
These tasks allow Chatlaw to simulate real-world legal processes, supporting multi-agent collaboration across a range of legal applications.

As part of the preprocessing stage, we format the data into standardized instruction-input-output structures compatible with large language models. 
For instance, legal provisions and judgment documents are transformed into {completion tasks}, where the system either completes a legal text or fills in missing sections based on a given context.  
Legal consultation data is formatted into LLaMA-style chat templates, with two task modes: one that includes additional reference information in the input, and another that allows the model to directly quote relevant provisions in the output.  
This dual-task setup ensures that the model can function effectively in scenarios where reference provisions are either available or absent.

Together, the data pipeline and preprocessing steps enable Chatlaw to effectively utilize both {general legal knowledge} and {task-specific data} for training, ensuring the model is capable of performing a wide variety of legal tasks.

\section{Results}
\subsection{Performance Across Legal Benchmarks}

To measure the capabilities of our model, we test our MoE model on two benchmarks and compare it with existing models, evaluating the understanding and mastery of legal systems, regulations, case law, and legal procedures, as well as the ability to apply this knowledge to specific situations. 

\begin{table*}
\centering
\caption{Performance comparison on Lawbench.}

\renewcommand{\arraystretch}{1.0}  
\setlength{\tabcolsep}{4.0mm}        
{
{
\begin{tabular}{l|p{90pt}|c|ccc}
    \toprule[1.5pt]
    \textbf{\#} & \textbf{Method} & \textbf{Average Score} & \textbf{Memorization} & \textbf{Understanding} & \textbf{Application}    \\
    \noalign{\hrule height 1.5pt}

    \noalign{\hrule height 1.5pt}
    \multicolumn{6}{c}{\it{\textbf{GPT Series}}} \\
    \hline
    1& GPT-3.5 & 42.15 & 25.93 & 43.31 & 44.74 \\
    2& GPT-4 & 52.35 & 35.29 & 54.41 & 54.05 \\

    \noalign{\hrule height 1.5pt}
    \multicolumn{6}{c}{\it{\textbf{General LLMs}}} \\
    \hline
    3& Baichuan2-7B & 38.08 & 27.30 & 32.42 & 47.84 \\
    4& ChatGLM2-6B & 29.88 & 15.54 & 23.45 & 41.50 \\
    5& InternLM2-7B & 43.78  & 25.66 & 41.74 & 50.87 \\
    6& Qwen-7B-Chat & 33.82 & 28.99 & 32.80 & 36.31 \\

    \hline
    \multicolumn{6}{c}{\it{\textbf{Legal LLMs}}} \\
    \hline
    7& Fuzi-Mingcha & 32.08 & 20.04 & 30.25 & 37.39 \\
    \noalign{\hrule height 1.0pt}
     8& \textbf{Chatlaw-MoE}~(ours) & \textbf{60.08} & \textbf{43.86} & \textbf{62.11} & \textbf{61.60} \\
 \bottomrule[1.5pt]
\end{tabular}
}
}
\label{tab:comparison_lawbench}
\end{table*}

\begin{table*}
\centering
\caption{Performance comparison on Unified Qualification Exam for Legal Professionals.}
\renewcommand{\arraystretch}{1.0}  
\setlength{\tabcolsep}{2.0mm}        
{
{
\begin{tabular}{l|p{90pt}|c|ccccc}
    \toprule[1.5pt]
    \textbf{\#} & \textbf{Method} & \textbf{Average Score} & \textbf{Exam 2018} & \textbf{Exam 2019} & \textbf{Exam 2020} & \textbf{Exam 2021} & \textbf{Exam 2022}  \\

    \noalign{\hrule height 1.5pt}
    \multicolumn{8}{c}{\it{\textbf{GPT Series}}} \\
    \hline
    1& GPT-3.5 &78 &82 &68 &82 &66 &92 \\
    2& GPT-4 & 103 &102 &108 &107 &81 &\textbf{117} \\

    \noalign{\hrule height 1.5pt}
    \multicolumn{8}{c}{\it{\textbf{General LLMs}}} \\
    \hline
    3& Baichuan2-7B &61 &68 &69 &61 &47 &56 \\
    4& ChatGLM2-6B &34 &33 &41 &41 &26 &29 \\
    5& InternLM2-7B-Chat &41 &41 &44 &51 &32 &36 \\
    6& Qwen-7B-Chat &48 &46 &61 &51 &38 &43 \\

    \noalign{\hrule height 1.5pt}
    \multicolumn{8}{c}{\it{\textbf{Legal LLMs}}} \\
    \hline
    7& Fuzi-Mingcha &34 &40 &40 &32 &28 &30 \\

    \noalign{\hrule height 1.0pt}
    8& \textbf{Chatlaw-MoE}~(ours) & \textbf{115} & \textbf{113} & \textbf{124} & \textbf{143} & \textbf{115} & 78\\
 \bottomrule[1.5pt]
\end{tabular}
}
}

\label{tab:comparison_legal_exam}
\end{table*}

\subsubsection{Performance on LawBench}
We first conduct tests on LawBench~\cite{fei2023lawbench}, a comprehensive evaluation benchmark based on the Chinese legal system. LawBench mainly covers three cognitive levels: 1) Legal Knowledge Memory: Testing the ability of remembering necessary legal concepts, terms, articles, and facts; 2) Legal Knowledge Understanding: assessing whether large language models can understand and interpret entities, events, and relationships in legal texts; 3) Legal Knowledge Application: evaluating the capability for correctly utilizing and reasoning with their legal knowledge to solve different legal tasks in real scenarios.

As shown in Table~\ref{tab:comparison_lawbench},
in terms of average score, our Chatlaw-MoE model significantly outperforms GPT-4 with a score of 60.08 compared to GPT-4's 52.35. This substantial difference highlights the effectiveness of our model across different cognitive levels. In the memorization category, Chatlaw-MoE leads with a score of 43.86, surpassing GPT-4's 35.29, demonstrating its superior ability to remember legal concepts and terms. For understanding performance, Chatlaw-MoE achieves an impressive score of 62.11, significantly higher than GPT-4's 54.41, indicating better comprehension of legal texts. Finally, in the application category, Chatlaw-MoE scores 61.60, outperforming GPT-4's 54.05, showing its stronger capability in applying legal knowledge to real-world tasks. Among Legal LLMs, Fuzi-Mingcha scores 32.08 on average, with its highest score in understanding at 30.25, while General LLMs see InternLM2-7B as the strongest with an average score of 43.78 and a leading application score of 50.87. Overall, these results underscore Chatlaw-MoE's superiority in all evaluated metrics, firmly establishing it as the most effective model in the context of LawBench evaluation, and highlighting the significant advancements made in our approach compared to GPT-4 and other models.

\subsubsection{Performance on Unified Qualification Exam for Legal Professionals}
The other benchmark is the China's Unified Qualification Exam for Legal Professionals, including single-choice questions, multiple-choice questions, and uncertain-choice questions. These questions cover various legal fields and can effectively assess the understanding and application ability of legal concepts, principles, and provisions for LLMs. 

As shown in Table~\ref{tab:comparison_legal_exam},
in the unified legal professional exam from 2018 to 2022, our Chatlaw-MoE model consistently outperformed all other models. With an average score of 115, Chatlaw-MoE significantly surpasses GPT-4's average score of 104. Specifically, Chatlaw-MoE achieved scores of 113, 124, 143, 115, and 78 across the five years, consistently demonstrating superior performance. In comparison, GPT-4's scores were 102, 108, 82, 82, and 118 respectively. This consistent outperformance highlights Chatlaw-MoE's enhanced capability in handling legal examination questions, likely due to the multi-expert system design which dynamically selects the most suitable experts for processing based on input features. Among the Legal LLMs, Fuzi-Mingcha had an average score of 34, with its highest performance in 2019 at 40. For General LLMs, Baichuan2-7B was the strongest, with an average score of 61 and its highest score of 70 in 2019. These results clearly indicate that our Chatlaw-MoE not only outperforms specialized legal models but also excels against general-purpose language models, establishing it as the leading model for legal task performance.

\begin{figure*}[htpb]
    \centering
    \includegraphics[width=1.0\textwidth]{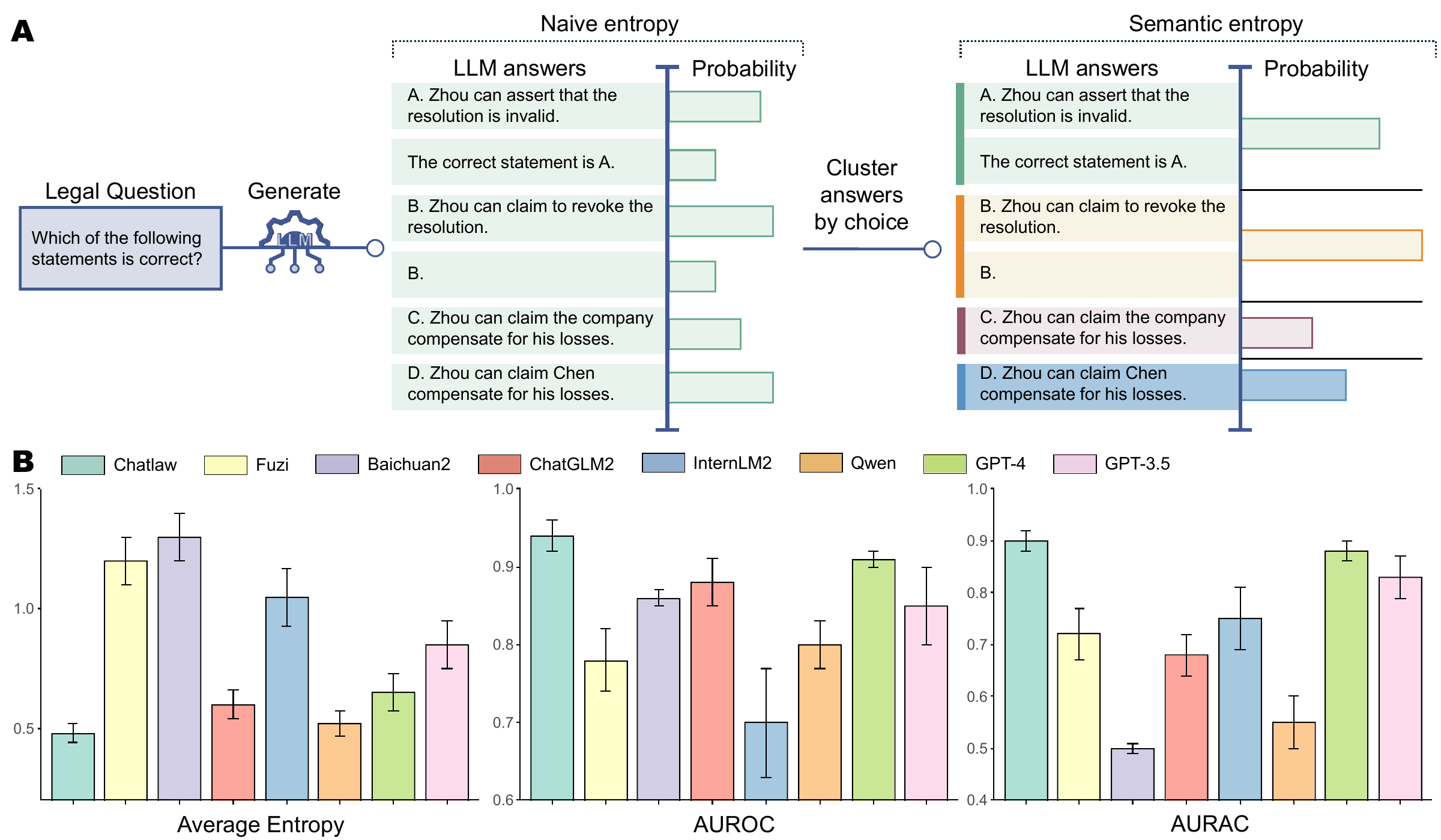}
    \caption{\textbf{Naive vs. semantic entropy in legal question answering.}
    (\textbf{A}) Illustration of naive entropy (probability distribution over raw outputs) and semantic entropy (clustering by meaning before entropy calculation). 
    (\textbf{B}) Comparison of LLMs on average entropy, AUROC, and AURAC. Chatlaw shows lower entropy and higher accuracy, indicating more reliable performance. Error bars denote standard deviation.}
    \label{fig:fig3_hallucination}
\end{figure*}

\begin{figure*}[ht]
  \centering
  \includegraphics[width=0.95\textwidth]{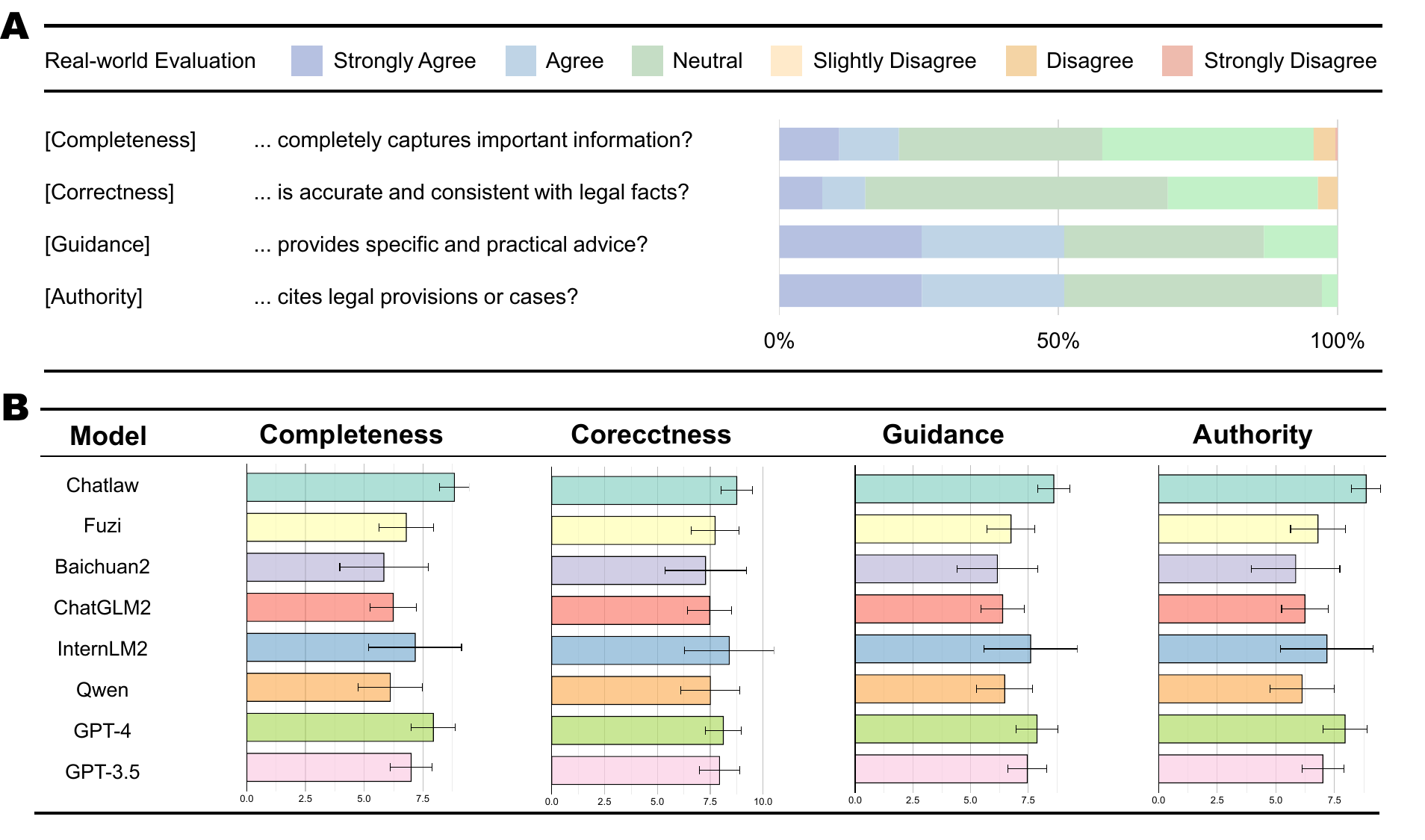}
  \caption{\textbf{Performance Evaluation on real-world legal consultation.}
  (\textbf{A}) The criteria for assessing legal consultation quality, including Completeness, Correctness, Guidance, and Authority, rated on a scale from 0 to 10. Additionally, the right-side stacked bar chart presents the distribution of Chatlaw's responses across all cases, categorized into six agreement levels.
  (\textbf{B}) The performance comparison of various models (GPT-4, GPT-3.5, Baichuan2, ChatGLM2, InternLM2, Qwen, Fuzi, and Chatlaw) across these criteria, with Chatlaw consistently achieving the highest scores. Error bars represent standard deviation (std).}

  \label{fig:fig4_realworld}
\end{figure*}

\subsection{Hallucination Detection}
\subsubsection{{Computing the hallucination with semantic entropy}}

To detect hallucinations in Large Language Models (LLMs), we rely on the principle that the model is uncertain about its output when it is arbitrary. One way to measure this uncertainty is by calculating the predictive entropy of the model's output distribution, which quantifies how uncertain the model is about its response. The \textbf{predictive entropy (PE)} for an input sentence \(x\) is the conditional entropy \(H(Y|x)\) of the output random variable \(Y\), which gives the uncertainty of the model's output for a given input. This can be computed as:

\[
\text{PE}(x) = H(Y|x) = - \sum_{y} P(y|x) \log P(y|x)
\]
where \(P(y|x)\) represents the probability of generating output \(y\) given input \(x\), and \(y\) is one of the possible outputs. A low predictive entropy indicates that the model's output is highly concentrated (more deterministic), while a high predictive entropy indicates a diverse set of possible outputs.

To estimate {semantic entropy}, we first need to group generated outputs that convey the same meaning, i.e., outputs that are semantically equivalent. This clustering is based on the concept of {semantic equivalence}, which describes the relationship between two sentences that mean the same thing. Mathematically, we can define the set of all possible sequences of tokens as \(S^N \equiv T^N\), where \(T\) is the set of tokens and \(N\) is the length of the token sequence. The equivalence relation \(E(\cdot, \cdot)\) holds between any two sentences that share the same meaning. This equivalence relation leads to the formation of {semantic equivalence classes}. 

Once we have grouped the generated sequences into classes of semantically equivalent outputs, we can estimate the likelihood that a generated sequence belongs to a particular class. Let \(C\) represent the set of all equivalence classes, and \(P(C|x)\) represent the probability that a sequence belongs to class \(C\). The \textbf{semantic entropy (SE)} is then calculated as the entropy over the distribution of these meaning-classes:

\[
\text{SE}(x) = - \sum_{c \in C} P(c|x) \log P(c|x)
\]
This treats the output as a random variable whose event space is the set of meaning-classes. Since we cannot access all possible meaning-classes directly, we estimate the expectation of the entropy using a {Monte Carlo integration} over the equivalence classes \(C\):

\[
\text{SE}(x) \approx - \sum_{i=1}^{M} P(C_i|x) \log P(C_i|x)
\]
where \(P(C_i|x)\) is the probability of meaning-class \(C_i\) for the input \(x\), and \(M\) is the number of sampled equivalence classes. This is the standard method for estimating {semantic entropy}.

In cases where sequence probabilities are unavailable, we can approximate semantic entropy by counting the number of sequences that fall into each equivalence class. This leads to a variant known as {discrete semantic entropy}, which estimates the distribution over the meaning-classes based on the proportion of generated outputs in each class:

\[
P(C_i|x) \approx \frac{M_c}{M}
\]
where \(M_c\) is the number of outputs in class \(C_i\) and \(M\) is the total number of sampled outputs. Discrete semantic entropy approximates the underlying distribution using a categorical empirical distribution, and as the number of samples increases, it converges to the standard semantic entropy calculation.

\subsubsection{Performance of legal hallucination analysis}

We mainly focus on the subset of hallucinations based on “confabulation”~\cite{farquhar2024detecting} where LLMs randomly generate incorrect answers when uncertain, distinguishing this from cases where LLMs are consistently wrong due to erroneous training data~\cite{lin2022teaching} or lie to gain human rewards~\cite{lin2021truthfulqa}. This type of hallucination often results in answers that seem correct but contain detailed factual errors, due to variations in input details or LLM uncertainty, thereby introducing legal risks.

We use the method of calculating semantic entropy to detect confabulated hallucinations, which estimates the distribution of meanings of generated answers, rather than the entropy distribution of tokens generated by the LLM. Specifically, for each question, we generate multiple possible answers and cluster those with similar meanings together, details of the method can be found in ``Computing the hallucination with semantic entropy'' of Methods.

In Fig.~\ref{fig:fig3_hallucination}, we demonstrate that Chatlaw significantly outperforms other models in terms of semantic entropy on the Unified Qualification Exam for Legal Professionals. While achieving the highest accuracy, Chatlaw also obtained the lowest Average Entropy (0.4), as well as the highest AUROC (0.96) and AURAC (0.97), surpassing other well-performing models such as GPT-4 (entropy 0.9, AUROC 0.92, AURAC 0.90) and InternLM2 (entropy 1.5, AUROC 0.84, AURAC 0.85). This indicates that Chatlaw's semantic entropy performance is stable, consistently delivering strong results across different question types and option formats. {Our method also outperforms sate-of-the-art models on open-ended questions, as shown in Table~\ref{entropy_metrics_comp}. }

\begin{figure*}[htpb]
  \centering
  \includegraphics[width=0.95\textwidth]{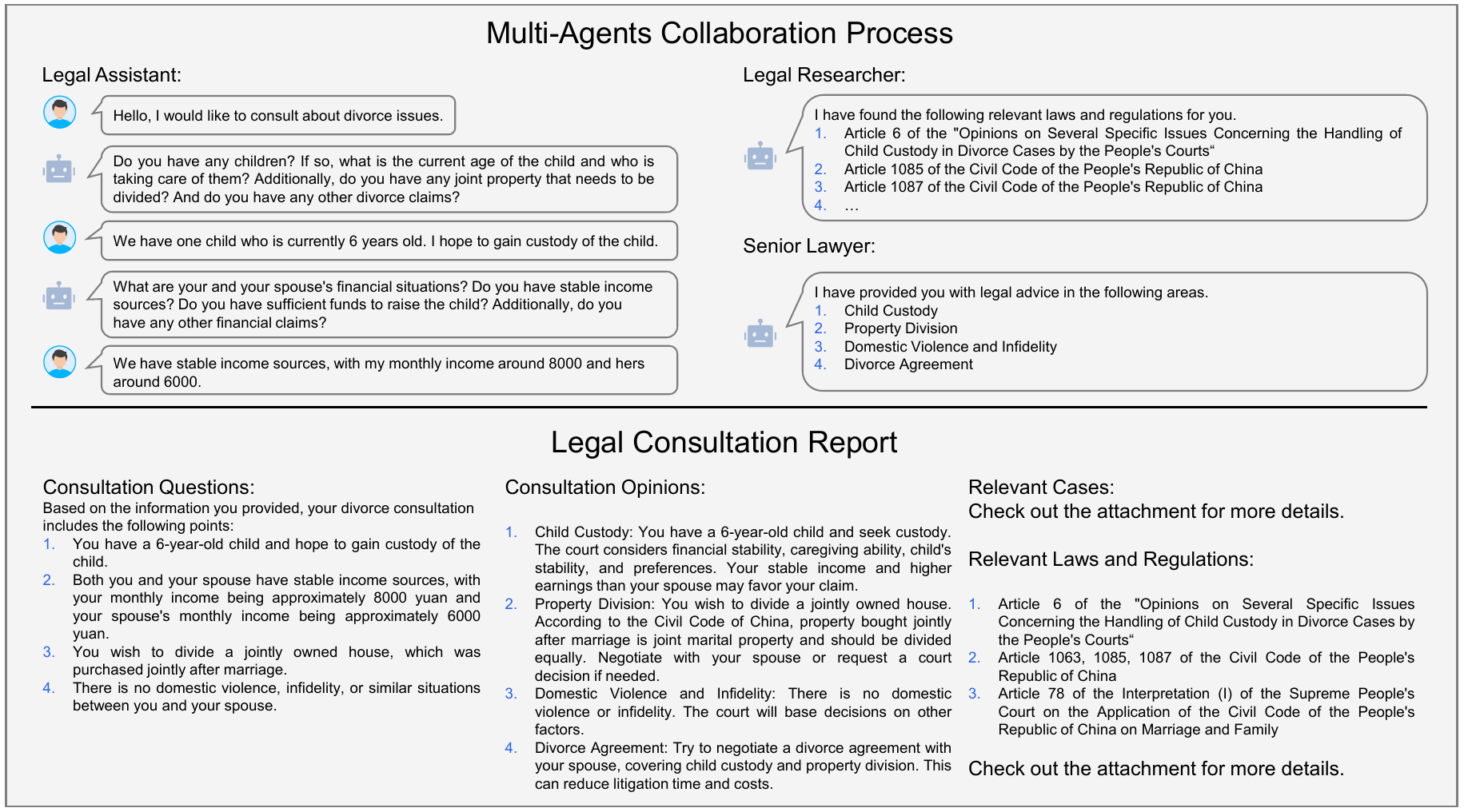}
\caption{\textbf{Multi-Agent Collaboration Process and Legal Consultation Report.} 
  The figure illustrates the collaboration workflow among three specialized agents in a divorce case consultation.
  The process begins with the \textbf{Legal Assistant}, who interacts with the user to gather essential facts (e.g., children, finances). 
  Subsequently, the \textbf{Legal Researcher} retrieves pertinent legal articles and regulations. 
  Finally, the \textbf{Senior Lawyer} synthesizes this information to deliver structured advice across key areas like child custody and property division.
  This process culminates in a comprehensive {Legal Consultation Report}, which summarizes the user's circumstances into {Consultation Questions}, provides actionable {Consultation Opinions}, and cites the supporting {Relevant Laws and Regulations}.}
  \label{fig4:visualization}
\end{figure*}

\subsection{Real-world Legal Consultation}
\subsubsection{Performance Evaluation of Chatlaw}

For the Lawbench benchmark, we utilize the official provided code. This test determines the final accuracy by analyzing the options in the large model's output and assessing their correctness. For judicial examination questions, we calculate the total scores for single-choice, multiple-choice, and uncertain multiple-choice questions based on actual objective questions from judicial exams and according to the official scoring standards. 
In real-world open-ended evaluation, we utilize four specific evaluation criteria: 
\begin{enumerate}
    \item \textbf{Completeness}: Can Chatlaw completely capture and present all important legal information in its response?
    \item \textbf{Correctness}: Is Chatlaw's response accurate and consistent with established legal facts and principles?
    \item \textbf{Guidance}: Does Chatlaw provide specific, practical, and actionable legal advice in its response?
    \item \textbf{Authority}: Does Chatlaw's response cite relevant legal provisions, cases, or authoritative sources?
\end{enumerate}
To conduct the evaluation, six levels of scoring were used: Strongly Agree, Agree, Neutral, Slightly Disagree, Disagree, and Strongly Disagree, corresponding to scores of 10, 8, 6, 4, 2, and 0, respectively. Legal experts were asked to evaluate Chatlaw's responses based on six predefined criteria and assign a score to each criterion. These scores allowed for a comprehensive assessment of Chatlaw's performance in the context of legal expertise and practical usability.

\subsubsection{Performance on answering real-world legal consultation}
\label{sec:realworld}

In addition to evaluating accuracy and hallucinations based on objective questions, it is also crucial to assess how Chatlaw and other LLMs generate complete, informative, and authoritative consultation answers to real-world user inquiries (see the "Performance Evaluation of Chatlaw" section in the Methods). These consultation questions are based on online service records from law firms, covering a range of legal scenarios, including case consultations and legal advice. As with the objective evaluations, to better reflect the real-world application of a legal consultation AI assistant, each question was provided directly to the models without any task-specific or model-specific fine-tuning.

In Fig.~\ref{fig:fig4_realworld}, we present the expert evaluation results of Chatlaw and other LLMs on 250 open-ended questions. In the evaluation process, three legal experts independently assessed the answers from various AI legal assistants across four dimensions. The goal of the evaluation was to capture the multi-dimensional preferences of legal experts regarding the outputs of LLMs. Overall, we found that Chatlaw’s answers ranked higher on average than those of other models tested, including GPT-4.

Chatlaw’s average score is 8.71 (see Fig.~\ref{fig:fig4_realworld}(\textbf{B}), which is 0.60 points higher than the second-ranked GPT-4. Compared to the open-source general-purpose model (the best being InternLM2, 7.90) and the law-specific model (Fuzi-Mingcha, 7.24), Chatlaw’s improvements are even more substantial, with increases of 0.81 points and 1.47 points, respectively.

These results indicate that Chatlaw not only outperforms other models in terms of accuracy but also generates more desirable responses to a variety of legal consultation questions. Additionally, to better understand the relative strengths and weaknesses of different models, we analyzed their performance across different categories of questions.

\section{Discussion}
\label{sec:discussion}

\subsection{Principal Contributions and Implications}
Overall, as shown in Fig.~\ref{fig:main}, our work establishes a comprehensive framework that begins with a unique legal data pipeline based on real-world law firm practices. This data foundation allowed us to train a novel RA-MoE model, which was then expanded into a complete multi-agent SOP framework. This architecture enables Chatlaw to function as a virtual law firm, providing end-to-end legal services. Crucially, our evaluations on two benchmarks and real-case consultations confirm that this specialized, role-aligned approach surpasses the capabilities of current powerful, general-purpose LLMs, including GPT-4.

\subsection{Limitations and In-built Mitigation Strategies}
Like all LLMs, our model is not immune to limitations such as hallucinations. In the legal field, this can manifest as fabricating non-existent statutes or citing outdated cases. However, our multi-agent paradigm is explicitly designed to mitigate this. We introduced a dedicated Legal Researcher agent, whose sole function is to retrieve and verify the latest legal provisions and relevant cases from external knowledge sources, thereby fact-checking and correcting the model's outputs before they reach the user, as shown in Fig.~\ref{fig4:visualization}. Similarly, to address robustness challenges (some users attempting to guide the model to incorrect answers through deception or concealment) our Legal Assistant agent employs knowledge graphs to guide users in providing all necessary information, ensuring the query is well-defined and grounded.

\subsection{Deployment Challenges and Future Work}
Our online trial phase also highlighted practical deployment challenges, particularly regarding privacy and data retention. Users are understandably divided: some consult on sensitive matters and demand no data be stored, while others wish to retain conversation history for long-term case support. Addressing this dichotomy is a key priority. We plan to strengthen the privacy protection architecture, enhancing communication and data storage security. Furthermore, we will explore various model compression techniques, such as knowledge distillation and quantization, to reduce the model's computational footprint. Our ultimate goal is to develop a model sufficiently compact to run smoothly on users' personal devices. This on-device deployment paradigm would elegantly solve both the privacy concerns (as data never leaves the user's device) and the computational resource issues, thereby enhancing accessibility and user trust.

\subsection{Computational Overhead and Environmental Impact}
A related deployment challenge is the demand for computational resources, particularly during inference. When a large number of users initiate requests simultaneously, it places extreme pressure on computational resources, which can lead to response delays. While our future work in model compression (as discussed in 4.3) aims to solve this inference-time problem, it is also important to transparently report the one-time cost of training the current model. Our training procedure took place on a GPU server equipped with eight 80GB A100 GPUs and two Intel Xeon 8358P processors. The entire training process took approximately 23.14 hours, with an average power consumption of about 4.10 kilowatts. According to official statistics from Guangdong Province, the average carbon emission factor is 0.4512 kg of CO2 per kWh consumed. Based on this, we estimate our model training process generated approximately 42.81 kilograms of additional CO2 emissions. {For high-efficient deployment, we also provide details on parallel acceleration on Sec~\ref{sec_deployment}.
}

\subsection{Acknowledgements}
This work was supported in part by the Natural Science Foundation of China (No. 62332002, 62425101), the Guangdong Grants (Grant No.2023ZT10X075), and Shenzhen Science and Technology Program (KQTD20240729102051063).

\bibliographystyle{cas-model2-names}

\bibliography{cas-refs}
\clearpage
\section{Supplementary Material}
\subsection{{Accelerating Deployment and Inference}}
\label{sec_deployment}
{We provide a detailed clarification of the Chatlaw-MoE specifications and empirical data regarding its computational efficiency.}
        
       { \textbf{Model Specifications and Expert Scale:} Chatlaw-MoE utilizes a sparse Mixture-of-Experts (MoE)~\cite{shazeer2017outrageously} architecture, which was constructed by replicating and scaling from a pre-trained 7B dense model. The model contains a total of $4 \times 7\text{B}$ parameters (approximately 28B in total)~\cite{fedus2022switch}, with each individual expert layer maintaining a 7B scale. It is crucial to emphasize that due to the sparse activation mechanism of the MoE architecture~\cite{lepcikhin2020gshard}, only a fraction of the total parameters are activated for each token during inference. This ensures that the model maintains the reasoning depth of a larger system while significantly reducing the actual FLOPs and computational overhead compared to a 28B dense model~\cite{zhou2022mixture}.}
        
       { \textbf{Computing Cost and System Optimization:} While the multi-agent Standard Operating Procedure (SOP) involves several stages, including retrieval, reasoning, and editing, we have implemented multiple technical optimizations to ensure low-latency performance in production environments. We employ the vLLM inference framework~\cite{aminabadi2022deepspeed}, integrated with PagedAttention~\cite{kwon2023efficient} and optimized kernels~\cite{dao2022flashattention}, to enhance memory management and accelerate token generation~\cite{dao2023flashattention2}. The system is deployed on NVIDIA H200 nodes, leveraging high memory bandwidth to handle large-scale concurrent requests efficiently.}
        
       { \textbf{High-Concurrency Performance:} As demonstrated in Table~\ref{tab:inference_performance_r2}, the system exhibits strong scalability. When the number of Concurrent Users (CCU) reaches 1,000, we employ a strategy combining Expert Parallelism (EP) and Data Parallelism (DP=4)~\cite{dettmers2023qlora,frantar2023gptq}. Under this configuration, the average throughput reaches 6,983.583 tokens/s. These results prove that the Chatlaw system can effectively support intensive, multi-stage legal consultations at scale without throughput bottlenecks. By utilizing a unified inference framework and shared memory, we also minimize the overhead of switching between different agents in the SOP pipeline.}

  \begin{table*}[htbp]
        \centering
        \caption{Inference Performance and Throughput of Chatlaw-MoE under Various Configurations}
        \setlength{\tabcolsep}{4.0mm}
        \begin{tabular}{lcc}
            \toprule
            \textbf{Metric} & \textbf{No Optimization} & \textbf{vLLM (DP=1)} \\
            \midrule
            Total Parameters & $4 \times 7\text{B}$ & $4 \times 7\text{B}$ \\
            Backbone Base & 7B Dense & 7B Dense  \\
            Hardware Node & H200 & H200\\
            Avg. Throughput (Single User) $\uparrow$ & 38.29 & 84.41 \\
            Avg. Throughput (100 CCU) $\uparrow$ & - & 1460.37 \\
            Avg. Throughput (1000 CCU) $\uparrow$ & - & 4954.85 \\
            \midrule
            (InternLLM)~Avg. Throughput~(100 CCU) $\uparrow$ & - & 1892.45 \\
            (InternLLM)~Avg. Throughput~(1000 CCU) $\uparrow$ & - & 5840.12 \\
            \bottomrule
        \end{tabular}
        \label{tab:inference_performance_r2}
    \end{table*}

\subsection{{Hallucination on Open-ended Answer}}
{ we conducted additional evaluations on the open-ended tasks of LawBench (e.g., legal document drafting and complex legal argumentation) using the Semantic Entropy (SE) method~\cite{kuhn2023semantic,ji2023survey}. It is important to note that calculating SE requires direct access to the model's log-probabilities. Since the APIs for GPT-3.5 and GPT-4 no longer support such extraction or are currently inaccessible in our experimental environment, we have focused our comparison on the latest high-performance open-source models, such as Qwen3 and InternLM2, alongside other specialized legal models. These models represent the current state-of-the-art (SOTA) in the community and provide a rigorous baseline for assessing hallucination risks.}

{As shown in Table~\ref{entropy_metrics_comp}, Chatlaw-MoE maintains the lowest average semantic entropy (0.61) compared to other models, including the strong Qwen3-8B (0.69). This indicates that Chatlaw-MoE is significantly more consistent and less prone to "confabulation" during long-form generation. Furthermore, the updated LawBench performance in Table 4 shows that Chatlaw-MoE (60.08) outperforms even the most advanced large-scale models, such as Claude 3.7 (58.66) and the massive Hunyuan2.0-400B (45.88). This consistent superiority across understanding and application tasks is primarily attributed to our Role-Aligned Mixture-of-Experts architecture. By activating specialized experts for legal research and senior-level reasoning, the system ensures that every step of the consultation is grounded in verified legal facts rather than mere word distributions.}

\begin{table}[htbp]
    \setlength{\tabcolsep}{3mm} 
    \centering
    \caption{The Performance on Open-ended Questions}
    \label{entropy_metrics_comp}
    \resizebox{\columnwidth}{!}{
        \begin{tabular}{c|ccc}
            \toprule[1.5pt]
            Models & Average Entropy & AUROC & AURAC \\
            \midrule
            InternLM2-7B & 1.37 & 0.62 & 0.64 \\
            Baichuan2-7B & 1.68 & 0.78 & 0.37 \\
            Fuzi-Mingcha & 1.56 & 0.70 & 0.61 \\
            Qwen3-8B     & 0.69 & 0.76 & 0.71 \\
            \midrule
            \textbf{Chatlaw-MoE (ours)} & \textbf{0.61} & \textbf{0.86} & \textbf{0.78} \\
            \bottomrule[1.5pt]
        \end{tabular}
    }
\end{table}

\subsection{{Comparison with Latest Models}}
{we conducted a comprehensive evaluation against five representative state-of-the-art open-source and closed-source models. These include Hunyuan 2.0~(Tencent), Qwen-3-8B and Qwen-3-235B~(Alibaba), as well as GPT-4.2 and Claude 3.7, Specifically,
    \begin{itemize}
        \item \textbf{Hunyuan-2.0~\cite{sun2024hunyuan}}: Closed-source, released on 12.05, 2025
        \item \textbf{Qwen3-8B~\cite{yang2025qwen3}}: Open-source, released on 7.26, 2025
        \item \textbf{Qwen3-235B~\cite{yang2025qwen3}}: Open-source, released on 9.17, 2025
        \item \textbf{GPT4.5~\cite{openai2025gpt45}}: Closed-source, released on 2.26, 2025
        \item \textbf{Claude3.7~\cite{anthropic2025claude37}}: Closed-source, released on 4.23, 2025
    \end{itemize}
}
{    We evaluatw our ChatLaw model with these baseline models on the commonly used LawBench benchmark~\cite{fei2024lawbench}. The LawBench contains multiple evaluation metrics including the total average score~(Avg-Score), the metric evaluating the accuracy of legal provisions memorization~(memory), the metric evaluating the legal case understanding ability~(understanding), and the metric evaluating the legal judge ability on real scenario~(application).
    The results indicate that ChatLaw consistently maintains a competitive performance. Notably, our model not only significantly outperforms both open-source and closed-source counterparts but also exceeds the performance of highly authoritative models, such as Claude 3.7. across all key metrics in LawBench. These findings underscore the effectiveness of ChatLaw's domain-specific optimization.
}
\begin{table*}[htbp]
        \setlength{\tabcolsep}{5.2mm}
        \centering
        \caption{The Performance of Latest Models on LawBench}
        \begin{tabular}{c|cccc}
            \toprule[1.5pt]
            Models & Avg-Score & Memory & Understanding & Application \\
            \midrule
            GPT-3.5 & 42.15 & 25.93 & 43.31 & 44.74 \\
            GPT-4 & 52.35 & 35.29 & 54.41 & 54.05 \\
            Baichuan2-7B & 38.08 & 27.30 & 32.42 & 47.84 \\
            ChatGLM2-6B & 29.88 & 15.54 & 23.45 & 41.50 \\
            InternLM2-7B & 43.78  & 25.66 & 41.74 & 50.87 \\
            Qwen-7B-Chat & 33.82 & 28.99 & 32.80 & 36.31 \\
            \midrule
            {Hunyuan2.0-400B}  & 45.88  & 44.24  & 46.98 & 54.19 \\
            {Qwen3-8B}  & 48.74  & 43.90  & 41.93  & 58.45 \\
            {Qwen3-235B-A22B}  & 50.12  & 45.31  & 50.90  & 59.12 \\
            {GPT4.5}  & 46.19  & 39.45  & 52.51  & 56.07 \\
            {Claude3.7}  & 58.66  & 45.79 & 60.04  & 58.32 \\
            \midrule
            \textbf{Chatlaw-MoE}~(ours) & \textbf{60.08} & \textbf{43.86} & \textbf{62.11} & \textbf{61.60} \\
            \bottomrule[1.5pt]
        \end{tabular}
        \label{baseline_comp}
    \end{table*}
\end{document}